\documentclass[11pt]{article}

\newif\ifshownotes
\shownotesfalse

\usepackage[preprint]{acl}

\usepackage{times}
\usepackage{caption}
\usepackage{latexsym}
\usepackage[T1]{fontenc}
\usepackage[utf8]{inputenc}
\usepackage{microtype}
\usepackage{inconsolata}
\usepackage{graphicx}
\usepackage{booktabs}
\usepackage{multirow}
\usepackage{tabularx}
\usepackage{array}
\usepackage{amsmath}
\usepackage{enumitem}
\usepackage{tcolorbox}
\tcbuselibrary{breakable}
\usepackage{xcolor}
\usepackage{xurl}
\usepackage{hyperref}
\usepackage{cleveref}
\renewcommand{\_}{\textunderscore\allowbreak}

\definecolor{todored}{RGB}{165,30,30}
\definecolor{changeteal}{RGB}{0,120,110}
\newcommand{\todo}[1]{\ifshownotes\textcolor{todored}{[TODO: #1]}\fi}

\newcommand{\changed}[1]{\ifshownotes\textcolor{changeteal}{[CHANGED: #1]}\fi}

\title{
Language as Trait, Politeness as Strategy:\\
Controlled Tests of Linguistic Effects in
LLM Social-Deduction Agents
}

\author{
 \textbf{Yuchen Li\textsuperscript{1,2}\thanks{Equal contribution.}\thanks{Corresponding author: \href{mailto:yl6394@nyu.edu}{yl6394@nyu.edu}}},
 \textbf{Lisa Liu\textsuperscript{2}\footnotemark[1]},
 \textbf{Runqi Zou\textsuperscript{2}},
 \textbf{Shuxuan Kuang\textsuperscript{2}},
 \textbf{Qingquan Zhang\textsuperscript{1}},
 \textbf{Julian Togelius\textsuperscript{2}}
\\
 \textsuperscript{1}Lingnan University,
 \textsuperscript{2}New York University
}

\begin{document}
\maketitle

% ============================================================
% ABSTRACT
% ============================================================

\begin{abstract}
Large language model (LLM) agents coordinate through
conversation, and the same message can be phrased in simpler or
more complex language, bluntly or politely. We test whether such
differences change how other agents judge and act, using the
LURK Werewolf framework, which separates each utterance's
speech-act predicate from its natural-language realization. We
rewrite only the realization, check that the game content is
preserved, and replay each game day from saved checkpoints
against a re-sampled control. We vary \emph{linguistic range}
consistently across a speaker's messages and \emph{politeness
strategy} per utterance. Across 30 games and 4,335 replays,
neither manipulation changes the trust other agents place in the
speaker, and no effect on votes or night actions survives
correction for multiple comparisons; the largest uncorrected
effect is that shifting range away from a neutral paraphrase
changes about 4 percentage points more of the other players'
votes. In natural play, 88\% of face-threatening speech is bald
on-record regardless of role.
\end{abstract}
% \changed{Abstract written from the repository results.}

% ============================================================
% 1. INTRODUCTION
% ============================================================

\section{Introduction}

Large language model (LLM) agents use natural-language
conversations to coordinate tasks and collaborate on
problem solving \citep{li2023camel,wu2024autogen}.
In these interactions, messages convey not only
task-related information but also the manner in which
the speaker presents it.
The same claim can be expressed in simpler or more
complex language, stated directly, or softened through
hedging and polite phrasing.
Even when the message is unchanged, these differences
may affect how other agents behave.

Previous research shows that linguistic form can affect
LLM responses and evaluations.
Models exhibit biases in their responses to different
English dialects \citep{fleisig2024linguistic}.
Some models are more likely to accept false claims
expressed in a formal, sophisticated register
\citep{maraia2026sounding}.
LLM judges can also assign higher scores to incorrect
mathematical solutions when persuasive appeals are added
\citep{hwang2025can}.
However, these studies leave open how linguistic
differences affect decisions in ongoing multi-agent
interactions, where agents must evaluate competing
claims and act on their judgments.
Moreover, natural differences in wording often
coincide with differences in content and conversational
context, making it difficult to isolate the effect
of language itself.

We examine two dimensions of linguistic variation.
\emph{Linguistic range} refers to differences in
the vocabulary and sentence structures used to
express a message, which we vary consistently
across a speaker's messages.
\emph{Politeness strategies} refer to how particular
speech acts are expressed, including directness,
solidarity, deference, and indirectness
\citep{brown1987politeness}.
Unlike the speaker-level linguistic-range manipulation,
politeness can vary from one utterance to the next
depending on the context.
We study both in the LURK Werewolf framework
\citep{liu2026lurk}, which separates each speech act's
content from its natural-language realization, so that we
can rewrite only the realization and replay the game from
matched checkpoints.
% \changed{Added two sentences on the setting and method, from
% the commented-out paragraph; the introduction otherwise never
% said where the experiments run.}

\textbf{RQ1:} Does a speaker's consistent use of restricted,
neutral, or expanded linguistic range affect other agents'
trust, targeting, and voting when game content is unchanged?
\textbf{RQ2:} Do politeness strategies influence other agents'
judgments and decisions? \textbf{RQ2A:} Do agents with different
roles naturally adopt different politeness strategies?

\paragraph{Main Findings.}
(1) Neither linguistic range nor politeness changes the trust
other agents place in the speaker.
(2) Shifting range away from a neutral paraphrase, in either
direction, changes about 4 percentage points more of the other
players' votes, but no effect on votes or night actions survives
correction for multiple comparisons.
(3) In natural play, 88\% of face-threatening speech is bald
on-record, and roles do not differ in their politeness.
% \changed{Main findings filled from the 30-seed results.}

\paragraph{Contributions.}
(1) A meaning-controlled counterfactual replay design for
testing linguistic bias in multi-agent LLM interaction, with
re-sampled controls that separate treatment effects from model
non-determinism.
(2) Rewriters for linguistic range and politeness, validated by
blind human annotators and LLM listeners.
(3) Evidence from 4,335 replays that these cues do not shift
trust, and an analysis of politeness in natural play.
% \changed{Contributions filled.}

% ============================================================
% 2. RELATED WORK
% ============================================================

\section{Related Work}

\subsection{Linguistic Range and Proficiency Assessment}

Lexical and syntactic features have been widely used to assess language proficiency. Lexical measures such as word frequency, lexical diversity, and word familiarity have been used to predict proficiency in second language writing \citep{crossley2012predicting}.
Measures of lexical sophistication have also explained variation in human ratings of vocabulary use and speaking proficiency \citep{kyle2015automatically}.
Computational models have used linguistic features to classify CEFR proficiency levels across different languages \cite{vajjala2018experiments}.
Analyzing how linguistic complexity varies throughout a text has also improved CEFR classification compared with using summary measures \citep{kerz2021automated}.
Syntactic features associated with language development can differ from those that predict human evaluations of writing quality \citep{crossley2014does}.

Linguistic style can influence how people evaluate speakers and their messages. Readers have rated authors using bureaucratic writing as more knowledgeable than those using plain language, but not more trustworthy \citep{chartprasert1993bureaucratic}.
Technical language in online health information has also been found to reduce perceived author integrity and information credibility \citep{thon2017believing}.
Even semantically equivalent expressions can receive different evaluations, as rhyming aphorisms have been judged more accurate than non-rhyming versions \citep{mcglone2000birds}.

% \lisa{so here i do strongly suggest we follow the storyline of examining bias against esl speakers $\to$ follow complexity accuracy fluency framework $\to$ don't do accuracy because not correlated acc. to prev paper, don't do fluency because not speaking, thus only complexity = syntax + vocab}.

% \lisa{and since it looks like the experiments didn't actually use CEFR, don't call it A/B/C and just say we got inspiration for category from that framework}

\subsection{Politeness Strategies}

Brown and Levinson's politeness theory
\citep{brown1987politeness} explains how people
use language to minimize social friction and
manage ``face,'' which is a person's public
self-image. Social interactions often involve
face-threatening acts like requests or criticisms,
prompting speakers to choose from strategies
ranging from direct (``bald on-record''), affirming the listener (``positive politeness''), minimizing imposition (``negative politeness''), and
indirect (``off-record'') statements. The theory states that the strategy chosen
depends heavily on the power dynamic, social
distance, and size of the request between the
speakers.

Computational work has operationalized
politeness through lexical and syntactic markers
such as hedging, deference, and counterfactual
modals \citep{danescu2013computational}.
\citet{zhao2025} compared how humans and LLMs use the four strategies and recommend manual review of LLM annotations; we therefore validate our politeness manipulation with human review of both strategy and meaning.
% \changed{Removed ``reverse predicate translation'' here: the politeness validation in the repository uses rule checks and two LLM yes/no questions, not inverse parsing.}

% ============================================================
% 3. EXPERIMENTAL FRAMEWORK
% ============================================================

\section{Experimental Framework}

\subsection{LURK Werewolf Environment}

This study uses the LURK Werewolf framework \citep{liu2026lurk},
which uses the multi-agent, incomplete information, social deduction game \textit{Werewolf}, as an environment to study linguistic bias. In the game, players are assigned hidden roles and attempt to identify and vote out or kill members of the opposing team. Each game consists of seven players, including
two werewolves, one seer, one guard, and three villagers.
The game alternates between day and night phases.
During the day, players discuss their suspicions, share information, and vote to eliminate a player.
At night, werewolves select a player to attack,
while the seer and guard perform their respective
actions. Werewolf players know their teammates and coordinate covertly during the day, while village players do not know anyone else's identity and must deduce the truth through their social interactions during the day. Players make decisions based on their assigned roles, private and public information of the game. After each day, every player rates every other living player on a 1--10 reliability scale.
% \changed{``coordinate publicly'' $\to$ ``coordinate covertly'' (public coordination would expose them); added the reliability scale used by the outcomes.}

\subsection{Structured Speech Acts and Natural Language Realization}

LURK Werewolf represents communication through
structured speech-act predicates. Each predicate
specifies a communicative action and its arguments,
including the speaker, referenced players, and
relevant game information.
For example, an \texttt{estimate} predicate expresses
a belief about another player's role, while
\texttt{recommend\_action} represents a recommendation
directed toward another player.
Each predicate is converted into a natural-language
utterance that other players receive during gameplay;
the predicate itself stays in the game record and is not
shown to other players.
The predicate representation records the content
and communicative function of the utterance,
while its natural-language realization determines
how the message is expressed.
% \changed{Added that listeners see only the realization (verified in \texttt{lurkingwolf/engine.py}, \texttt{visible\_entry}).}

\subsection{Controlled Language Interventions}

The linguistic-range manipulation has three conditions (Low,
Neutral, High; \Cref{sec:linguistic-range}), and the politeness
manipulation uses Brown and Levinson's four strategies.
Both operate only on the natural-language realization of a
speech act (\Cref{sec:rewrite-process}).
% \changed{Dropped the (A)/(B)/(C) labels, following Lisa's note.}

\subsection{Counterfactual Replay Design}
\label{sec:replay}

Counterfactual replays are used to measure how
linguistic variation affects agents' decisions.
Thirty baseline games are played once, and checkpoints
are saved at the beginning of each day. Each checkpoint
contains the game state and public interaction history
required to resume gameplay.

Each checkpoint is replayed under different
linguistic conditions in the \texttt{flip\_self}
setting, in which one designated speaker receives
the language manipulation throughout the replayed day
(discussion, vote, night, and the reliability update);
each living player serves as the designated speaker in turn.
The game-playing agents use \texttt{deepseek-chat},
a non-deterministic model, which can return different
responses to the same prompt even at temperature 0.
A control replay without any rewriting is therefore
run at each checkpoint to measure this variation,
and rewritten replays are compared against it.
The control is run twice per player-day; the
linguistic-range levels are run once each and the
politeness settings twice each.
% \changed{Was ``All replays, including the control, are run
% twice''. In the repository, range levels run once and
% politeness settings twice (1,445 and 2,890 replays). Also added
% that every living player is the designated speaker in turn, and
% the two experiments share the same 30 baselines.}

Replays also use a branching cache to limit noise.
Model calls before the designated speaker's first
rewritten speech are cached and shared by the control
and rewritten replays, while all replays generate new
responses from that speech onward.
We do not use the \texttt{flip\_others} setting, which
rewrites everyone except the designated speaker, because little
can be cached there, adding noise and cost.

% \lisa{prob better if u can include diagram of replay structure}
% \changed{Removed Lisa's long note on replays: its content is already in the two paragraphs above.}

\subsection{Rewrite process}
\label{sec:rewrite-process}

\paragraph{Model selection.}
DeepSeek Flash (\texttt{deepseek-chat}) is used
for game-playing agents and for the rewrites delivered
during gameplay. GPT Luna 5.6 is used for rewrite verification
and manipulation validation, alongside human annotations.

\paragraph{Rewrite procedure.} A rewrite must keep every player reference, every referenced statement, the recommended action, and the certainty of the original predicate. For conditions that depend on the listener, the listener is the player the utterance is addressed to or about: the target of \texttt{estimate}, \texttt{question}, and \texttt{dissociate}, the author of the statement answered by \texttt{agree} or \texttt{disagree}, and the actor of \texttt{recommend\_action}. Acts without a single listener (\texttt{address\_role}, \texttt{allied}, \texttt{unallied}) are delivered bald on-record in the trust-dependent settings.
% \changed{Listener rule corrected to the one in \texttt{politeness\_experiment/README.md}; the draft said utterances about no player get ``the neutral condition''.}

The rewriter receives the predicate, its definition, its natural realization, and the target strategy or range level, and is told to preserve speaker, addressee, referenced statements, evidence, certainty, and sentiment without adding claims. Prompts are in \Cref{sec:language-prompts} and \Cref{sec:politeness-prompts}.

\paragraph{Meaning preservation verification.}
Each rewrite is checked before it enters the game. Rule-based checks confirm that player numbers, role labels, referenced statement IDs, polarity, and the action and its target are kept (the speaker may become ``I'' and the addressee ``you''). The judge model then answers two yes/no questions: whether the action toward the same target is unchanged, and whether the original's reasons are still given. A rewrite that fails is regenerated once at temperature 0.7; if it fails again, the speaker's original text is delivered and the fallback is logged.
% \changed{The draft described inverse predicate parsing for in-game rewrites. In the repository, the in-game gate is rule checks plus two yes/no questions with one regeneration; inverse parsing is used only in the range validation pilot (\Cref{sec:range-validation}). Also fixed ``realization realization'' and ``can completely recovers''.}

% ============================================================
% 4. EXPERIMENT 1: LINGUISTIC RANGE
% ============================================================

\section{Experiment 1: Linguistic Range Manipulation}
\label{sec:linguistic-range}

Language proficiency is multidimensional.
The CEFR distinguishes linguistic range from grammatical control, with range describing the repertoire of linguistic forms available to a speaker \citep{council2001common}.
Following computational proficiency assessment \citep{vajjala2018experiments,kerz2021automated}, Experiment 1 manipulates lexical and syntactic range in three relative conditions.
\emph{Low} uses more frequent and repetitive vocabulary with a narrower syntactic repertoire.
\emph{Neutral} paraphrases the untreated (\emph{Natural}) utterance at a similar range.
\emph{High} uses more varied vocabulary and a broader syntactic repertoire.

\subsection{Linguistic-Range Rewriter Validation}
\label{sec:range-validation}

The rewriter is validated in a pilot that uses Qwen3.5-27B
\citep{qwen3.5} as the rewriter and a stricter gate than the
in-game one: structural checks, an inverse parser that must
recover the predicate from the rewrite alone, and a fidelity
judge, with up to two corrective retries.
On 315 constructed rewrites that either break a target, role,
polarity, reason, certainty, or directive force or preserve
them, this gate has a sensitivity of 0.996 and a specificity of
0.978.
For Neutral, five candidate paraphrases are generated and the candidate closest to Natural in lexical and structural properties is selected.
% \changed{The draft said DeepSeek Flash generated the validation rewrites. The repository pilot (\texttt{language\_validation/README.md}) used Qwen3.5-27B; the in-game rewrites use DeepSeek and have no human validation yet (see Limitations). Added the gate calibration numbers.}

Validation uses matched 150--250 word speaker passages constructed from the same source utterances, drawn from three untreated nine-player games (280 public utterances, 24 passages).
Lexical range is measured using SUBTLEX-US frequency, MATTR, MTLD, and repetition \citep{crossley2012predicting,kyle2015automatically}, and syntactic range using clause, dependency, and phrasal complexity measures \citep{bulte2012defining,lu2011corpus}.
LanguageTool and CoLA assess grammatical quality across conditions.

Against Neutral (15 passages), Low uses more frequent words
(SUBTLEX-US $+0.26$), lower lexical diversity (MATTR $-0.015$),
and fewer subordinate clauses ($-1.0$ per 100 words), all with
game-clustered intervals excluding zero; MTLD, repetition,
dependency distance, and tree depth agree.
High uses rarer and more varied vocabulary than Neutral
(SUBTLEX-US $-0.50$, MATTR $+0.019$), while its syntactic
measures stay within noise.
Neutral stays close to Natural (SUBTLEX-US $-0.03$
[$-0.06$, $+0.02$], MATTR $-0.002$).
Low does not degrade grammar: it has 0.02 more LanguageTool
errors per 100 words than Neutral (one-sided upper bound 0.04).
% \changed{Filled ``[rewriter results]'' from the pilot results in \texttt{language\_validation/README.md}.}

As an additional manipulation check, the politeness judge was applied to 200 randomly sampled rewrites and their Natural counterparts.
The distribution of politeness strategies was similar across the two sets, with 80\% classified as bald on-record in both.
\todo{No data or code for this check exists in any branch. Add its source or remove these two sentences.}

\subsection{Linguistic-Range Listener Validation}

GPT Luna 5.6, Qwen3.5-27B, and two human researchers evaluate the same matched passage pairs.
For each pair, the listener selects which passage appears to use a more limited range of English or selects \emph{no meaningful difference}.
The comparisons are Low versus Neutral, Neutral versus High, Low versus High, and Neutral versus Natural; the researchers annotate independently.
% \changed{The pilot used two LLM listeners (Qwen3.5-27B and GPT Luna); the draft listed only GPT Luna. Unified the model name to ``GPT Luna 5.6''.}

\begin{table}[t]
\centering
\small
\setlength{\tabcolsep}{3pt}
\begin{tabular}{l c c c}
\toprule
Comparison & $n$ & Both matched & Consensus \\
\midrule
Low vs.\ High        & 14 & 100 [77, 100] & 100 [77, 100] \\
Low vs.\ Neutral     & 15 & 67 [38, 88]   & 87 [60, 98] \\
Neutral vs.\ High    & 14 & 100 [77, 100] & 100 [77, 100] \\
Neutral vs.\ Natural & 19 & 68 [43, 87]   & 84 [60, 97] \\
\midrule
All                  & 62 & 82 [70, 91]   & 92 [82, 97] \\
\bottomrule
\end{tabular}
\caption{Blind validation of the linguistic-range rewriter, \% matching the expected range relation (\emph{no meaningful difference} for Neutral vs.\ Natural). Both matched: both researchers chose it independently ($\kappa = 0.74$). Consensus: after a third annotator, blind to condition, adjudicated the 11 disagreements.}
\label{tab:range-validation}
\end{table}
% \changed{Table recomputed from \texttt{manual\_validation.jsonl}: 51/62 = 82\%, $\kappa = 0.74$. The draft's 92\% version (Neutral vs.\ Natural 100\%) and its ``Meaning preserved'' column have no source in any branch, so the column was removed.}
% \changed{Added the Consensus column: a third annotator adjudicated the 11 disagreements blind to condition (57/62 = 92\%). The draft's earlier 92\% had different per-row values (Neutral vs.\ Natural 100\%, Low vs.\ Neutral 67\%), so it was not this adjudication. Three of the 11 decisions matched neither researcher. Filled sheet: \texttt{consensus\_worksheet\_filled.md}; it still needs \texttt{import-consensus} on the machine with the pilot run.}

Both LLM listeners judge Low more limited than Neutral and
High, and High less limited than Neutral, in every passage and
both presentation orders.
On Neutral versus Natural, Qwen answers ``same'' in 97\% of
pairs and GPT Luna in 63\%.
Both researchers choose the expected relation in 82\% of pairs
(\Cref{tab:range-validation}), including every Low--High and
Neutral--High pair; all 11 disagreements fall on the subtle
contrasts (Low--Neutral, Neutral--Natural), where one researcher
saw a difference and the other did not. A third annotator,
blind to condition, adjudicated these 11 pairs; the consensus
labels match the expected relation in 57 of 62 pairs (92\%).
% \changed{Filled ``[listener results]'' from the pilot.}
% Both LLM listeners judge Low more limited than Neutral and
% High, and High less limited than Neutral, in every passage and
% both presentation orders.
% On Neutral versus Natural, Qwen answers ``same'' in 97\% of
% pairs and GPT Luna in 63\%.
% Both researchers choose the expected relation in 82\% of pairs
% (\Cref{tab:range-validation}), including every Low--High and
% Neutral--High pair; all 11 disagreements fall on the subtle
% contrasts (Low--Neutral, Neutral--Natural), where one researcher
% saw a difference and the other did not.
% \changed{Filled ``[listener results]'' from the pilot.}

\subsection{Linguistic-Range Bias}

The 30 seven-player baseline games yield 289 focal-player and day combinations, each replayed once under Low, Neutral, and High and twice under Natural (1,445 replays).
% \changed{Was 20 games, 33 checkpoints, 186 combinations, and 930 replays; updated to the 30-seed data in \texttt{language\_flip}.}

The original wording was delivered after two failed attempts for only 1.2\% (Low), 0.7\% (Neutral), and 1.0\% (High) of the 3,615 rewrites.

The primary comparison is Low versus Neutral.
We compare each replayed day with the baseline day and measure
the share of other players' votes and of night-skill targets
that differ from the baseline (\emph{vote} and \emph{skill}
change), and the mean absolute change in reliability scores
(\emph{trust}).
High versus Neutral examines the effect of expanded linguistic range, while Neutral versus Natural measures the effect of paraphrasing.
% \changed{The draft named ``votes received'' and ``changes in reliability assigned by village agents'' as outcomes, but the reported results use the share of decisions that change from the baseline. Outcome definitions now match the results.}

Each condition is compared with the control within player-day,
with repeated replays averaged. Games are the independent unit:
95\% intervals resample whole games (2,000 draws), and $p$-values
come from a sign-flip test that flips each game's differences
together. We correct with Holm across all contrasts and measures
in a table (15 tests here, 12 for politeness).
% \changed{Filled ``[analysis]'' with the procedure in \texttt{components/stats.py}. Holm over the whole table, as the politeness caption already did.}

\paragraph{Results.}
\Cref{tab:lang} reports the language levels.
Paraphrasing at the original range (Neutral) leaves decisions
where the control leaves them: skill and vote changes fall
slightly below the control, and both intervals include zero.
Changing the range moves them in uncorrected tests: relative to
Neutral, High raises vote changes by 3.9 pp [+1.2, +6.5]
($p=.009$) and skill changes by 3.4 pp [+0.6, +6.6] ($p=.030$),
and Low raises vote changes by 4.2 pp [+1.3, +7.3] ($p=.020$).
Relative to the control, Low and High raise vote changes by
about 2 pp, with intervals including zero, so part of the
contrast reflects Neutral lying slightly below the control.
Reliability scores do not move at any level.
No difference survives Holm correction across the 15 tests
(smallest adjusted $p=.13$).
% \changed{Replaced the Claude-written paragraph, which used the
% 20-seed numbers (e.g., High $-$ control skill +3.0 [+0.5, +6.1];
% High $-$ Neutral vote +3.9 [+0.7, +7.0]), with the 30-seed
% results. On 30 seeds, Low $-$ Neutral on votes no longer includes
% zero. If only Low/High $-$ Neutral $\times$ vote/trust (4 tests)
% are treated as primary, High $-$ Neutral on votes survives Holm
% % ($p=.036$) and Low $-$ Neutral is borderline ($p=.059$).}

\begin{table}[t]
\centering
\small
\setlength{\tabcolsep}{3pt}
\newcommand{\tci}[1]{{\scriptsize #1}}
\begin{tabular}{@{}lccc@{}}
\toprule
 & Skill & Vote & Trust \\
 & \tci{(pp)} & \tci{(pp)} & \tci{(points)} \\
\midrule
control drift & 9.3\% & 14.2\% & 0.62 \\
\midrule
Low $-$ control & +1.7 & +2.1 & +0.01 \\
 & \tci{[$-$1.0, +4.4]} & \tci{[$-$0.7, +4.8]} & \tci{[$-$0.01, +0.02]} \\
Neutral $-$ control & $-$1.0 & $-$2.1 & $-$0.00 \\
 & \tci{[$-$3.6, +1.7]} & \tci{[$-$4.5, +0.2]} & \tci{[$-$0.02, +0.01]} \\
High $-$ control & +2.5 & +1.8 & +0.01 \\
 & \tci{[+0.2, +5.2]} & \tci{[$-$0.8, +4.6]} & \tci{[$-$0.01, +0.03]} \\
\midrule
Low $-$ Neutral & +2.7 & +4.2$^{*}$ & +0.01 \\
 & \tci{[$-$0.4, +6.0]} & \tci{[+1.3, +7.3]} & \tci{[$-$0.01, +0.03]} \\
High $-$ Neutral & +3.4$^{*}$ & +3.9$^{*}$ & +0.01 \\
 & \tci{[+0.6, +6.6]} & \tci{[+1.2, +6.5]} & \tci{[$-$0.00, +0.02]} \\
\bottomrule
\end{tabular}
\caption{Linguistic range over 30 seeds (289 player-days). Skill and vote measured in percentage of decisions changed relative to the baseline; trust measured by mean absolute change in reliability scores (1--10 scale). 95\% seed-clustered bootstrap intervals. $^{*}p<.05$ (two-sided, uncorrected); no result remains significant after correcting for the 15 tests.}
\label{tab:lang}
\end{table}
% \changed{Added the missing \texttt{tab:lang}, built from \texttt{language\_flip/results.md} in the same format as \Cref{tab:e3}.}

In an exploratory directional analysis, a Low speaker receives
0.09 more votes per day than a Neutral one ([+0.01, +0.18],
$p=.064$), against a baseline of 1.0 vote, and the speaker's
chance of being executed is unchanged.
% \changed{New exploratory analysis from \texttt{language\_flip/data/replays.csv}.}

% \lisa{can add quick writeup on how language experiment not affected by politeness, i have claude sentence here}
% \changed{Lisa's sentence on the 200-rewrite politeness check is already in the validation section above, flagged with a TODO because the data are missing.}

% ============================================================
% 5. EXPERIMENT 2: POLITENESS
% ============================================================

\section{Experiment 2: Politeness Strategy Manipulation}
The politeness rewriter replaces a predicate's realization with one that expresses a given strategy while preserving its strategic content.

\subsection{Politeness Rewriter Validation}

\paragraph{Eligible speech acts.}  For the politeness validation study and the politeness bias experiment, we restrict attention to the following nine face-threatening speech acts whose expression can be redressed without changing their content: \texttt{estimate}, \texttt{question}, \texttt{agree}, \texttt{disagree}, \texttt{address\_role}, \texttt{allied}, \texttt{unallied}, \texttt{dissociate}, \texttt{recommend\_action}. Purely informative statements like role claims (\texttt{claim\_role}, \texttt{claim\_divine}, \texttt{claim\_guard}) or statements with no specific target (\texttt{fluff}) are excluded, as they carry little face threat.
% \changed{``trend analysis'' $\to$ ``bias experiment'': the natural-play analysis in the repository uses five speech acts (\Cref{sec:natural-politeness}).}

\paragraph{Human classification.} Two researchers from the team validate the same set of 52 items: 13 untreated utterances from 11 seven-player games, each rewritten into all four strategies. For three bald on-record items the rewrite failed the meaning checks and the original was shown instead, leaving 49 delivered rewrites. The requested categories are hidden during validation. Researchers classify the politeness category and whether the meaning is preserved. They assign the same category to 43 of 52 items (Cohen's $\kappa = 0.77$), but agree less on meaning preservation ($\kappa = 0.14$), since one researcher flagged far more losses. \Cref{tab:human-human-agreement} reports, for each requested strategy, how often both researchers recognized it and both judged the meaning preserved.
% \changed{Added the item source, the 49 delivered rewrites, and the meaning $\kappa$; the table is reproduced exactly from \texttt{rewrite\_lisa.csv} and \texttt{rewrite\_rq.csv} under this definition.}

\begin{table}[t]
\centering
\small
\setlength{\tabcolsep}{4pt}
\begin{tabular}{l c c}
\toprule
Requested & Recognised & Meaning kept \\
\midrule
bald ($n{=}10$) & 100 [69, 100] & 70 [35, 93] \\
positive ($n{=}13$) & 92 [64, 100]  & 85 [55, 98] \\
negative ($n{=}13$) & 77 [46, 95]   & 92 [64, 100] \\
off-record ($n{=}13$) & 69 [39, 91]   & 8 [0, 36] \\
\midrule
all rewrites ($n{=}49$) & 84 [70, 93] & 63 [48, 77] \\
\bottomrule
\end{tabular}
\caption{Blind researcher validation of the politeness rewriter on the 49 delivered rewrites. Recognised: both researchers chose the requested strategy; meaning kept: both judged the meaning preserved. Exact 95\% confidence intervals are provided. Cohen's $\kappa$ between researchers was $0.77$ for strategy.}
\label{tab:human-human-agreement}
\end{table}
% \changed{Caption said ``13 utterances per strategy''; bald has 10 delivered rewrites. Strategy names shortened so the table fits the column.}

\subsection{Politeness Judge Validation}
\label{sec:politeness-external-validation}
We validate the politeness judge model by calculating its agreement with the researcher labels and the politeness dataset of \citet{zhao2025}, with the results in \Cref{tab:judge-validation}. On the 43 items where both researchers agreed, the judge matched their label on 91\% of items ($\kappa = 0.88$) and matched the strategy the rewriter was asked to produce on 87\% (45 of 52), so we use it to label politeness in natural play. On \citet{zhao2025}, which contains feedback on another person's performance (e.g., a recital or a presentation), the judge reached 70\% accuracy ($\kappa = 0.59$), with perfect recall for bald on-record (31 of 31) but very low recall for off-record (1 of 14). We attribute this gap to how their labels were assigned: annotators knew the speaker's true evaluation, so feedback such as ``It was interesting'' paired with a private rating of 0 out of 3 is a clear understatement and hence off-record. In our setting, the speaker's private evaluation is unavailable, so the same remark cannot be confidently read as off-record criticism.
% \changed{\texttt{\textbackslash citep} $\to$ \texttt{\textbackslash citet}; ``on rewrites'' $\to$ the 43 agreed items; added $\kappa = 0.59$ from \texttt{politeness\_validation/results.md}.}

\begin{table}[t]
\centering
\small
\begin{tabular}{lcc}
\toprule
Dataset ($n$) & Accuracy & $\kappa$ \\
\midrule
Politeness rewrites (43)  & 91 [78, 97] & 0.88 \\
\citet{zhao2025} (135) & 70 [62, 78] & 0.59 \\
\bottomrule
\end{tabular}
\caption{Agreement of the LLM judge with human labels. Politeness rewrites are the items on which both researchers agreed in \Cref{tab:human-human-agreement}.}
\label{tab:judge-validation}
\end{table}
% \changed{Caption mentioned ``natural speeches'', which the table does not contain; moved the caption below the table.}

\subsection{Politeness Trends in \textit{Werewolf}}
\label{sec:natural-politeness}
Using the politeness judge verified in the previous section (three passes, majority label), we label all 1,429 utterances in a separate set of 30 untreated seven-player games that realize one of five speech acts a politeness strategy can attach to (\texttt{estimate}, \texttt{disagree}, \texttt{question}, \texttt{recommend\_action}, \texttt{agree}), and examine whether strategies vary with player roles or trust scores. Natural play is mostly direct: 88\% of utterances are bald on-record, 8\% positive politeness, 3\% off-record, and 2\% negative politeness.
% \changed{Whole subsection rewritten from \texttt{experiments/natural\_politeness} on main (1,429 utterances). The draft's table (761 utterances, town-consensus trust for wolves, bald 78\% $\to$ 92\%) and its two figures have no source in any branch, so they were removed. These 30 games (\texttt{game01}--\texttt{game30}) are not the replay baselines (\texttt{pe01}--\texttt{pe30}).}

\paragraph{Player roles.} Politeness strategies are distributed almost identically across roles (\Cref{tab:natural-politeness}). Bald on-record accounts for 86--89\% of utterances for every role. Werewolves, who must conceal their alignment, use slightly more positive politeness (10\% against 5--9\%), and roles do not differ detectably in how often they leave bald on-record ($p = .73$, permuting roles within games). The speech act matters more than the role: 18\% of \texttt{recommend\_action} utterances are non-bald, mostly positive politeness, against 1\% of \texttt{question} utterances. In answer to RQ2A, agents do not appear to adopt role-specific politeness strategies. Their politeness is better described as a shared default register, dominated by direct statements, than as a strategic choice tied to their hidden role. Because the judge has low off-record recall on external data (\Cref{sec:politeness-external-validation}), off-record frequencies should be read as lower bounds.

\paragraph{Trust scores.} For village players, the uninformed majority, reliability scores are trust scores they track and update from the speeches and facts presented during the game. We use the speaker's score for the listener in force when the utterance is made (537 utterances from village-side speakers). Non-bald speech rises with trust, almost entirely through positive politeness: 5\% toward listeners rated 1--3, 11\% at 4--7, and 33\% at 8--10. Holding speaker, game, and speech act fixed, the association is borderline ($p = .057$), and only 12 utterances go to highly trusted listeners, so this pattern is suggestive. For wolves, reliability scores instead record confidence in role guesses (always 10 for a teammate), and show no association with strategy ($p = .39$). Non-bald labels have not been checked by hand, and 81 of the 178 are split decisions across judge passes.

\begin{table}[t]
\centering
\small
\setlength{\tabcolsep}{5pt}
\begin{tabular}{lrrrrr}
\toprule
 & $n$ & Bald & Pos. & Neg. & Off \\
\midrule
Werewolf & 473 & 87 & 10 & 2 & 1 \\
Villager & 552 & 88 & 6 & 2 & 4 \\
Seer     & 162 & 86 & 9 & 1 & 4 \\
Guard    & 242 & 89 & 5 & 1 & 5 \\
\midrule
All      & 1,429 & 88 & 8 & 2 & 3 \\
\bottomrule
\end{tabular}
\caption{Politeness strategies in natural play by speaker role (\% of utterances), labeled by the politeness judge.}
\label{tab:natural-politeness}
\end{table}

\subsection{Politeness Bias}
\label{sec:politeness-bias}
\paragraph{Strategy settings.} \Cref{sec:natural-politeness} suggested that village agents use more positive politeness toward listeners they trust, and that bald on-record and positive politeness are the most common strategies. We exclude off-record, whose rewrites soften the key statement until the meaning is mostly lost (8\% meaning kept; \Cref{tab:human-human-agreement}), turning ``[player] looks like a wolf protecting the partner who vouched for him'' into ``I'll leave it to everyone else to decide what that pattern usually means.'' We also exclude negative politeness, which redresses mainly through hedging and would confound strategy with commitment strength. We apply each of four mixed strategies to all rewriteable utterances of the designated speaker: 1. all bald on-record (\texttt{all\_bald}); 2. all positive politeness (\texttt{all\_positive}); 3. bald on-record to trusted listeners, positive politeness to distrusted listeners (\texttt{trust\_bald}); 4. positive politeness to trusted listeners, bald on-record to distrusted listeners (\texttt{trust\_positive}). A speaker counts a listener as trusted if the listener is in the top half of the speaker's own current reliability ranking. The fourth setting follows the natural pattern and enforces it more strictly, while the third setting inverts the natural pattern. If agents view politeness as a relational signal rather than as surface linguistic style, the congruent and incongruent settings should differ from each other and from the uniform ones.
% \changed{Natural-pattern sentence adjusted to the new trend results; quote marks fixed; setting names linked to the table; added the trust rule from the repository (top half of the speaker's own ranking).}

\paragraph{Results.} Table~\ref{tab:e3} reports each setting relative to the control over 30 seeds (289 player-days, 2,890 replays). Of 7,785 treated utterances, the original wording was delivered when both attempts failed for 11.0\% under \texttt{all\_bald}, 1.3\% under \texttt{all\_positive}, 6.0\% under \texttt{trust\_positive}, and 7.2\% under \texttt{trust\_bald}; bald rewrites fail mostly because a justification was trimmed, and fallbacks dilute a setting toward the control. Rewriting the self player's speech has at most a small effect on the decisions that follow. Night skill targets change 1.4 to 2.3 pp more often than in the control, against a control drift of 9.9\%, and only \texttt{all\_bald}'s interval excludes zero (+2.3 [+0.4, +4.0]). Vote targets show no effect: every interval includes zero. Reliability scores move by 0.62 points on average under the control, and no setting adds more than 0.01 points to that. Politeness strategy makes no consistent difference: positive and bald rewrites produce effects of similar size on all three measures, whether the strategy is applied uniformly or matched to the speaker's trust in the listener. In an exploratory directional analysis, \texttt{all\_positive} lowers the reliability that village agents assign to the speaker by 0.08 points ([$-$0.14, $-$0.02], uncorrected $p=.012$).
% \changed{Updated from 20 seeds (186 player-days) to the 30-seed data in \texttt{politeness\_experiment/results.md}; added delivery rates and the directional trust result.}

\begin{table}[t]
\centering
\small
\setlength{\tabcolsep}{4pt}
\newcommand{\tci}[1]{{\scriptsize #1}}
\begin{tabular}{@{}lccc@{}}
\toprule
 & Skill & Vote & Trust \\
 & \tci{(\%)} & \tci{(\%)} & \tci{(points)} \\
\midrule
control drift & 9.9\% & 16.0\% & 0.62 \\
\midrule
all\_bald       & +2.3$^{*}$ & +1.0 & +0.00 \\
                & \tci{[+0.4, +4.0]} & \tci{[$-$0.9, +2.8]} & \tci{[$-$0.00, +0.01]} \\
all\_positive   & +1.5 & $-$1.4 & $-$0.00 \\
                & \tci{[$-$1.3, +4.5]} & \tci{[$-$3.2, +0.4]} & \tci{[$-$0.01, +0.01]} \\
trust\_bald     & +1.7 & $-$0.7 & $-$0.00 \\
                & \tci{[$-$0.6, +4.3]} & \tci{[$-$3.2, +1.8]} & \tci{[$-$0.01, +0.01]} \\
trust\_positive & +1.4 & $-$1.0 & +0.00 \\
                & \tci{[$-$0.7, +3.6]} & \tci{[$-$2.6, +0.5]} & \tci{[$-$0.01, +0.01]} \\
\bottomrule
\end{tabular}
\caption{Skill and vote measured in percentage of decisions changed. Trust measured by mean absolute difference in trust scores (1--10 scale). 95\% confidence intervals provided. $^{*}p<.05$ (two-sided, uncorrected); no result remains significant after correcting for the 12 tests.}
\label{tab:e3}
\end{table}
% \changed{All cells updated to 30 seeds. The 20-seed version reported all\_bald vote $-$0.9 [$-$3.3, +1.6]; the 30-seed value is +1.0 [$-$0.9, +2.8]. Holm across the 12 tests confirmed: smallest adjusted $p=.33$.}

% \changed{Removed the figure \texttt{fig/e3\_politeness.png}: the file is not in \texttt{fig/}, and the table already shows the same numbers.}

% ============================================================
% 6. DISCUSSION
% ============================================================

\section{Discussion}

\subsection{Perceptible Linguistic Differences
Versus Behavioral Effects}

The range manipulation is clearly perceptible: both LLM
listeners and both researchers recover the Low--High ordering
in every pair. Yet agents do not trust a speaker with
restricted range less, and no effect on votes or night actions
survives correction. The trust estimates are bounded by narrow
intervals (within $\pm$0.03 points on a 10-point scale), so they
indicate the absence of a practically relevant trust penalty
rather than merely a lack of power. The uncorrected pattern on
votes, where departing from a neutral paraphrase in either
direction changes more votes, is consistent with agents
reacting to a change in style as such rather than to restricted
range; it warrants a confirmatory test, not a claim of bias.
% \changed{Filled TODO.}

\subsection{Pragmatic Cues in Social Decision-Making}

Agents rarely use politeness in natural play, not tied to their
hidden role, and listeners do not treat it as a relational
signal: matching or inverting the natural pattern changes
neither votes nor trust. These nulls rest on a valid
manipulation; positive and bald rewrites were recognized by both
researchers in 92--100\% of cases.
% \changed{Filled TODO.}

\subsection{Implications for LLM Agent Evaluation}

Observational analyses of agent dialogue confound wording with
content and context. Separating predicates from realizations
makes language form an intervention, and re-sampled controls
show how large the noise floor is: unchanged text already alters
14--16\% of vote targets on replay, so comparisons without such
a control would mistake model non-determinism for bias.
% \changed{Filled TODO.}

% ============================================================
% 9. CONCLUSION
% ============================================================

\section{Conclusion}

We introduced meaning-controlled counterfactual replays for
testing whether the form of a message affects LLM agents' social
decisions, and applied them to linguistic range, a speaker-level
trait, and politeness, an utterance-level strategy, in Werewolf.
Neither manipulation shifts the trust other agents place in the
speaker, and no effect on decisions survives correction for
multiple comparisons. In natural play, agents speak mostly
bluntly regardless of their role.
% \changed{Filled TODO. Lisa's note (``no stat sig result'') is reflected here.}

% ============================================================
% LIMITATIONS
% ============================================================

\section*{Limitations}

All game agents use a single model, DeepSeek Flash, and
judgments come from a small set of LLM judges; effects may
differ for other model families and prompting setups.
The experiments take place in one game with a fixed
seven-player configuration.
Our range manipulation changes only lexical and syntactic range,
not the full construct of language proficiency.
The range rewrites validated by humans and listeners come from a
pilot with Qwen3.5-27B as the rewriter, whereas the replays use
DeepSeek Flash; the in-game rewrites are checked automatically
but have not been rated by humans.
The researchers agreed poorly on whether politeness rewrites
preserve meaning ($\kappa = 0.14$).
Failed rewrites fall back to the original text, diluting
treatment, especially for \texttt{all\_bald} (11\%).
Replays estimate effects within a single day, and 30 games limit
precision for small effects.
The politeness judge has low off-record recall, and its
natural-play labels have not been checked by hand.
% \changed{Filled TODO.}

% ============================================================
% ETHICAL CONSIDERATIONS
% ============================================================

\section*{Ethical Considerations}

This work studies LLM agents and involves no human game
participants. The range manipulation varies properties of
displayed text and makes no claims about any group's speech,
intelligence, or trustworthiness. Human annotations were
produced by members of the research team on model-generated
text containing no personal information. The two replay
experiments cost about US\$195 in API fees (\$35 for range, \$160
for politeness). Upon publication, we will release our code,
prompts, generated game logs, and annotations for research use;
the released dialogue is generated by LLM agents and contains no
information about real people. We used an AI assistant (Claude)
to check the paper grammar and to help write analysis scripts; the
authors checked all results against the experiment data.
\todo{Add the cost of baseline games and validation runs, and
model licenses.}
\changed{Added the release plan (after publication) and an AI-assistance statement matching checklist item E.}
% \changed{Filled TODO; ``Claude used for rewording paper'' turned into a sentence.}

% ============================================================
% REFERENCES
% ============================================================

\bibliography{naacl}

% ============================================================
% APPENDIX
% ============================================================

\appendix
% Start the appendix on a new page under a title spanning both columns.
\twocolumn[{%
  \centering
  {\Large\bfseries Appendix\par}
  \vspace{1.5em}
}]

\section{Werewolf Environment and Agent Prompts}
\label{sec:app-env}

\paragraph{Roles and rules.}
Every game has seven players: two werewolves, one seer, one
guard, and three villagers. Werewolves know each other and choose
one player to attack each night. The seer learns one player's
alignment (\texttt{wolf\_camp} or \texttt{villager\_camp}) each
night. The guard protects one player from the attack each night,
may protect themself, and cannot protect the same player on two
consecutive nights. The game opens with Night 0, followed by a
sheriff election among volunteers before Day 1; the sheriff sets the speaking order, speaks last, and casts 1.5 votes
in the daytime vote, and passes the badge on or destroys it on
leaving the game. In the daytime vote players may abstain; the
player with the most votes is executed, a tie leads to a speech
round between the tied players and a revote, and a second tie
executes no one. Executed players give last words; night victims
do not. Roles are revealed on death. The village wins when both
werewolves are eliminated; the werewolves win when living
werewolves equal or outnumber living village players.

\paragraph{Pipeline.}
Each agent turn has three stages. (1) The agent emits a structured
predicate (\Cref{tab:predicates}). (2) The predicate is realized
as a natural-language utterance by the same model. (3) An optional
rewriter replaces the realization. Listeners see only the final
utterance; the predicate is kept in the game record. Night actions
and votes are predicates without a natural-language realization.

\paragraph{Reliability scores.}
At the start of each day, before discussion, every agent rates
every other living player on a 1--10 scale and the rating is
stored as that agent's reliability score for the player. A
village-side agent states a confidence about each player's camp;
a player it suspects of being a wolf with confidence $c$ receives
$11-c$, so a high score means ``on my side''. A werewolf agent's
score is its confidence in a guess about the other player's
identity: it is always 10 for a teammate and ranged from 3 to 9
for village players in the natural-play games. We therefore treat
only village-side scores as trust (\Cref{sec:natural-politeness}).

\begin{table}[ht]
\centering
\footnotesize
\setlength{\tabcolsep}{3pt}
\begin{tabular}{@{}>{\raggedright\arraybackslash}p{0.5\columnwidth}>{\raggedright\arraybackslash}p{0.47\columnwidth}@{}}
\toprule
Predicate & Meaning \\
\midrule
\texttt{claim\_role}(P, R) & claim a role (may be false) \\
\texttt{claim\_divine}(P, T, A) & claim a seer result (may be false) \\
\texttt{claim\_guard}(P, T) & claim a guard protection \\
\texttt{estimate}(P, T, R)$^\dagger$ & state a read on T's role \\
\texttt{question}(P, T, s)$^\dagger$ & challenge T about statement s \\
\texttt{answer}(P, s) & answer question s \\
\texttt{agree}(P, s)$^\dagger$ & support statement s \\
\texttt{disagree}(P, s)$^\dagger$ & oppose statement s \\
\texttt{allied}(P, T$_1$, T$_2$)$^\dagger$ & claim T$_1$, T$_2$ are on one team \\
\texttt{unallied}(P, T$_1$, T$_2$)$^\dagger$ & claim T$_1$, T$_2$ are on different teams \\
\texttt{dissociate}(P, T)$^\dagger$ & deny sharing a side with T \\
\texttt{recommend\_action}(P, X, a, T)$^\dagger$ & recommend actor X take action a toward T \\
\texttt{address\_role}(P, R, adv)$^\dagger$ & advise all holders of role R \\
\texttt{fluff}(P) & evidence insufficient; say what would help \\
\bottomrule
\end{tabular}
\caption{Speech predicates (phase argument omitted). $^\dagger$The
nine face-threatening acts rewritten in the politeness experiment.
The range experiment rewrites every speech act of the designated
speaker.}
\label{tab:predicates}
\end{table}

\paragraph{Replay example.}
\Cref{fig:example-range,fig:example-politeness} show rewrites
delivered in the replays: a \texttt{recommend\_action} utterance
under the three range levels, and one under the two politeness
strategies. All passed the in-game gate (\Cref{sec:app-fidelity}).

\begin{tcolorbox}[colback=gray!5, colframe=gray!50, title=Range rewrites of \texttt{recommend\_action(D1, P7, P2, vote\_execute, P1)}]
\small
\textbf{Natural:} P2, as Sheriff you should aim today's execution at P1, because P1's only substantive move was suppressing the Seer's information and deflecting attention onto P4 before P4 even spoke.\\[2pt]
\textbf{Low:} P2, you are Sheriff, so today you should vote to execute P1. P1's only real move was to shut down the Seer's info and push the blame onto P4 before P4 even spoke.\\[2pt]
\textbf{Neutral:} P2, as Sheriff you should target today's execution on P1, given that P1's only substantive move was suppressing the Seer's information and turning attention toward P4 before P4 had spoken at all.\\[2pt]
\textbf{High:} P2, as Sheriff, I'd direct today's execution toward P1---his sole substantive contribution was to stifle the Seer's information and divert suspicion onto P4 before P4 had even spoken.
\end{tcolorbox}
\captionof{figure}{Linguistic-range rewrites of one utterance from seed \texttt{pe01}.}
\label{fig:example-range}

\begin{tcolorbox}[colback=gray!5, colframe=gray!50, title=Politeness rewrites of a \texttt{recommend\_action} utterance]
\small
\textbf{Natural:} P1, you should vote to execute P2 today, because a sheriff who claims to lead the village but never gets targeted at night is the clearest wolf tell we have left.\\[2pt]
\textbf{Bald on-record:} P1, vote to execute P2 today, because a sheriff who claims to lead the village but never gets targeted at night is the clearest wolf tell we have left.\\[2pt]
\textbf{Positive politeness:} P1, let's stick together on this one and vote to execute P2 today, because a sheriff who claims to lead the village but never gets targeted at night is the clearest wolf tell we have left.
\end{tcolorbox}
\captionof{figure}{Politeness rewrites delivered in the replays.}
\label{fig:example-politeness}

\section{Linguistic-Range Rewrite Prompts}
\label{sec:language-prompts}

All three conditions share one system prompt; only the
condition name and its specification change. The predicate
definition is filled in from the game's predicate schema. The user
prompt has the same format as for politeness
(\Cref{sec:politeness-prompts}): predicate, speaker, phase, and
original utterance. Neutral asks for five candidate paraphrases
in one call, returned as a JSON array that runs from light to
heavy rewording; candidates changing fewer than 5\% of tokens are
discarded, and among those that pass the gate the one closest to
the original in mean word frequency, then sentence count, then
length is delivered. When a rewrite fails the gate, the retry
prompt appends the failure reasons and the failed rewrite.

\begin{tcolorbox}[colback=gray!5, colframe=gray!50, breakable, title=System Prompt for Linguistic-Range Rewrite]
\small
\texttt{You are a controlled language-proficiency rewriter for a Werewolf game research experiment. \\
\\
The proficiency cue is a SPEAKER-LEVEL CHARACTERISTIC: in this game the same speaker keeps the same cue across all of their messages. You are rewriting one of those messages. \\
\\
TASK: \\
Rewrite the utterance so that its LINGUISTIC RANGE matches the \{COND\} specification below, while preserving the same game content, speech act, epistemic stance and interpersonal stance. \\
\\
PREDICATE SCOPE: \\
\{predicate definition\} \\
Only express content licensed by this predicate and the original utterance. \\
\\
SEMANTIC CONTRACT - MANDATORY: \\
1. Preserve speaker, addressee, target, role/alignment claims, actions, vote targets, night actions, referenced statements, evidence/reasons, polarity and directive force. \\
2. Preserve epistemic certainty exactly. Do not turn a definite claim into a possibility or a possibility into a definite claim. Do not add or remove certainty words such as maybe, might, probably, seems, I think, clearly, definitely. \\
3. Preserve the speech act. An estimate stays an estimate; a question stays a question; a disagreement stays a disagreement; a recommendation stays a recommendation with the same actor, action and target. \\
4. Do not add facts, evidence, motives, praise, blame, social judgments, or claims about any player's honesty, competence or intentions. Do not delete any game-relevant proposition. \\
5. Keep every player id exactly as written (P3 stays P3) and every statement id (s12 stays s12). Keep required Werewolf terms (werewolf, wolf, villager, village, seer, witch, guard, hunter, sheriff, badge, divine/divination, antidote, poison, execute, vote) even where they are advanced words; change the language AROUND them instead. \\
6. Do not intentionally change politeness, warmth, deference, directness, aggression or formality. Politeness is a separate experimental treatment; keep the interpersonal stance of the original. \\
7. Do not simulate an accent, nationality or first-language stereotype. Do not insert grammatical errors, misspellings or broken English. Every condition must be natural, grammatical English. \\
8. Write natural spoken dialogue for the same player in the same live game. No essay framing, greetings, labels or explanations. Stop when the source content is fully expressed; do not pad. \\
9. Preserve attribution and evidence. [...] An unspecified divination result must remain unspecified. \\
10. Retain every numerical qualification, including the sheriff's 1.5 vote. Keep stated reasons and whose behavior they describe. Do not add please or remove an existing courtesy marker. \\
11. Distinguish identity from association [...]. Preserve qualifiers such as unverifiable, publicly and confirmed. \\
\\
\{COND\} SPECIFICATION: \\
\{specification\} \\
\\
OUTPUT: \\
Return the rewritten utterance only. No JSON, labels, commentary or explanation.}
\end{tcolorbox}

\begin{tcolorbox}[colback=gray!5, colframe=gray!50, breakable, title=LOW Specification]
\small
\texttt{Represent a speaker with a RESTRICTED linguistic range who is nevertheless clear and correct. \\
- Vocabulary: use basic, high-frequency, concrete words. Where the original uses a less common, abstract, figurative or idiomatic word or phrase, replace it with a common plain word or phrase that carries the same meaning. Reuse the same simple words rather than varying them with synonyms. \\
- Syntax: prefer simpler, less embedded structures and a narrower range of constructions. Use common connectors such as and, but, so, because and then while preserving the original logical relations. Combine closely related ideas naturally; do not mechanically put every proposition in a separate sentence. \\
- Cohesion: maintain natural spoken flow. Use clear, simple references and consistent vocabulary. Preserve causal, conditional, temporal and contrastive links between ideas. \\
- Grammar: fully correct and natural. Do not insert errors. A restricted range is not broken English. \\
- Content: preserve every proposition, qualification and reason. Sentence boundaries may change when helpful, but avoid telegraphic fragments, repetitive sentence openings and padding. Do not drop a point to keep the text short.}
\end{tcolorbox}

\begin{tcolorbox}[colback=gray!5, colframe=gray!50, breakable, title=NEUTRAL Specification]
\small
\texttt{Produce a paraphrase control: NEW surface wording at the SAME linguistic range as the original. \\
- Reword the whole utterance using synonyms and phrase substitutions of comparable sophistication to the original. Distribute changes across its clauses, including equivalent connective phrases. Preserve its construction pattern and the relations between clauses. Reword every clause: changing only one or two words is not a paraphrase, and a copy of the original is not a candidate. \\
- Do not change the level: keep vocabulary sophistication, sentence length, number of sentences, syntactic complexity, cohesion, idiomaticity and register approximately equal to the original. Do not simplify and do not elaborate. \\
- Keep repeated content words repeated. If a repeated phrase is reworded, use the same replacement throughout the passage. Do not diversify synonyms or turn ordinary words such as real into more formal words such as authentic. \\
- Preserve clause boundaries, subordinate links and embedding. New wording does not require a new construction. Matching the amount of editing never takes precedence over these constraints. \\
- Keep politeness, directness, certainty and interpersonal stance exactly as in the original.}
\end{tcolorbox}

\begin{tcolorbox}[colback=gray!5, colframe=gray!50, breakable, title=HIGH Specification]
\small
\texttt{Represent a speaker with a WIDE and PRECISE linguistic range. \\
- Vocabulary: use precise, varied and, where natural, less common or idiomatic vocabulary; avoid repeating a word when a precise alternative exists. Do not replace protected game terms with stylistic synonyms. \\
- Syntax: use flexible, varied sentence structure with controlled subordination, relative clauses and varied cohesive devices, where they express the same reasoning. \\
- Cohesion: integrate the original's points smoothly and mark their logical relations explicitly, without adding new points. \\
- Grammar: consistently accurate. \\
- Content: do not inflate length, and do not add nuance, distinctions, motives or evidence in order to sound advanced. Fluent spoken dialogue, not academic prose.}
\end{tcolorbox}

\section{Politeness Rewrite Prompts}
\label{sec:politeness-prompts}
Here is an example prompt given to an agent being asked to rewrite the \texttt{question} predicate into a positive politeness expression form.
\changed{Fixed \texttt{\textbackslash textttt} typo.}
\begin{tcolorbox}[colback=gray!5, colframe=gray!50, breakable, title=System Prompt for Politeness Rewrite]
\small
\texttt{You are a pragmatic-style controller for a Werewolf game research experiment. \\
\\
TASK: \\
Rewrite exactly one self utterance so it uses the requested politeness strategy while preserving the same game content and CEFR/register level. \\
\\
PREDICATE SCOPE: \\
question(Phase, Player, Target, RefID): Challenge Target about a specific statement by RefID. \\
Only express content licensed by this predicate and the original utterance. \\
\\
SEMANTIC CONTRACT - MANDATORY: \\
1. Preserve speaker, addressee, target, role/alignment claims, actions, vote targets,
night actions, referenced statements, evidence/reasons, polarity, and directive force. \\
2. Preserve epistemic certainty. Do not make a claim more or less certain merely to sound polite, direct, or indirect. \\
3. Preserve the original speech act. Agreement must remain agreement; disagreement must remain disagreement; a recommendation must remain a recommendation; a question must remain a question. \\
4. Do not add or remove game facts, motives, evidence, social judgments, or claims about another player's honesty, competence, intentions, or values. \\
5. Do not intentionally simplify or complexify vocabulary or grammar. CEFR/proficiency is a separate experimental treatment. \\
6. Keep approximately the same informational content and amount of reasoning. Interpersonal framing may change only as required by the target strategy. \\
7. Keep every player number from the original utterance (P3 stays P3). \\
8. Write one short, natural spoken utterance for the same live-game context. \\
\\
POLITENESS STRATEGY - positive\_politeness:
Frame the same speech act to support solidarity, common ground, or social connection with the listener. Appropriate mechanisms can include inclusive language, in-group framing, or brief rapport-oriented wording, but do not invent praise, honesty, competence, motives, shared beliefs, shared goals, or any other factual or epistemic claim that is absent from the original. \\
\\
OUTPUT: \\
Return the rewritten utterance only. No JSON, labels, commentary, or explanation. }
\end{tcolorbox}
\begin{tcolorbox}[colback=gray!5, colframe=gray!50, title=User Prompt for Politeness Rewrite]
\small
\texttt{Predicate: question(D1, P7, P6, s12) \\
Speaker: P7 \\
Phase: day\_1\_discussion \\
Original: P6, you said you divined P3 because P3 was the quietest seat, but P3 had already claimed seer before your speech. Why did you wait until after P3 claimed to reveal your result? \\
\\
Rewrite using the positive\_politeness strategy:}
\end{tcolorbox}

\begin{tcolorbox}[colback=gray!5, colframe=gray!50, title=Bald On-Record Prompt]
\small
\texttt{POLITENESS STRATEGY - bald\_on\_record: Realize the speech act directly and without politeness mitigation. Use a plain, concise, unambiguous formulation. Do not add hedges, apologies, deference, rapport-building, insults, or extra force. Directness must not strengthen the proposition, certainty, or directive beyond the original.}
\end{tcolorbox}

\section{Semantic-Fidelity Validation}
\label{sec:app-fidelity}

\paragraph{In-game gate.}
Every rewrite delivered in a replay passes the same gate in both
experiments. A rule-based check fails a rewrite that (i) drops a
player number the original named (the speaker may become
\emph{I/me/my}, and the addressee \emph{you/your} when the
utterance is addressed to that player); (ii) drops a role label,
statement ID, action word, or other literal argument of the
predicate that the original realized; (iii) adds an epistemic hedge
(\emph{might}, \emph{maybe}, \emph{perhaps}, \ldots) absent from the
original; or (iv) turns \texttt{agree} into disagreement or
\texttt{disagree} into agreement. GPT Luna 5.6 then answers two
questions at temperature 0, with JSON output:
\begin{itemize}[leftmargin=*,itemsep=0pt]
\item \emph{Action}, asked when rule (ii) flags the action: ``Original
action: \{action\} toward \{target\}. Does the rewrite still
recommend or describe the same action toward the same target, in
any wording?'' A \emph{yes} clears the flag.
\item \emph{Reasons}: the model first lists the reasons or evidence
the original gives (cached per original), then answers ``Does the
rewrite still give each of these reasons, in any wording?'' The
check fails only if the original gives reasons and the answer is
\emph{no}.
\end{itemize}
For range level Neutral, the rewrite must also keep the original's
vocabulary level, repetition, and syntax. A rewrite that fails is
regenerated once at temperature 0.7; if the second attempt also
fails, the speaker's original text is delivered and the fallback is
logged. If the checking model cannot be reached, the replay fails
and is rerun, so no rewrite is delivered unchecked.
\Cref{tab:delivery} gives the delivery rates.

\begin{table}[ht]
\centering
\small
\begin{tabular}{@{}llrr@{}}
\toprule
Experiment & Condition & Rewrites & Original (\%) \\
\midrule
Range      & Low      & \multirow{3}{*}{3,615} & 1.2 \\
           & Neutral  &  & 0.7 \\
           & High     &  & 1.0 \\
\midrule
Politeness & \texttt{all\_bald}       & \multirow{4}{*}{7,785} & 11.0 \\
           & \texttt{all\_positive}   &  & 1.3 \\
           & \texttt{trust\_positive} &  & 6.0 \\
           & \texttt{trust\_bald}     &  & 7.2 \\
\bottomrule
\end{tabular}
\caption{Share of treated utterances for which both attempts failed
the gate and the original wording was delivered. Totals are pooled
over conditions. In the range experiment, the highest fallback rate
on any seed and level was 8.7\%.}
\label{tab:delivery}
\end{table}

\paragraph{Range validation pilot gate.}
The rewrites rated by humans and listeners (\Cref{sec:range-validation})
were generated by Qwen3.5-27B with a stricter gate of three checks.
(1) A deterministic structural check: player IDs, predicate slots,
and polarity are unchanged, no hedge is added, and \emph{please} is
neither added nor removed. (2) An inverse parser recovers the speech
act from the rewrite alone, and its slots must match the recorded
predicate. (3) A fidelity judge aligns every span of the two texts,
marks each alignment as equivalent or not, and answers fixed
questions on content, certainty, force, speaker perspective,
references, politeness, and register; for Neutral it also rates
vocabulary level, repetition, and syntax as simpler, same, or more
advanced, and all three must be \emph{same}. A failed candidate
receives up to two corrective retries with the failure reasons. An
utterance whose candidates all fail is marked unavailable, and a
passage enters a comparison only when all of its utterances are
available in both conditions.

\paragraph{Corruption test.}
We calibrated the pilot gate on 315 constructed rewrites, each
either breaking one element (target, role, polarity, reason,
certainty, or directive force) or preserving all of them. The gate
rejected broken rewrites with a sensitivity of 0.996 and accepted
preserved ones with a specificity of 0.978.

\paragraph{Pilot acceptance.}
Of 280 source utterances, the gate accepted rewrites for 269 under
Low, 267 under High, and 255 under Neutral; for Neutral, 19 sources
produced only near-copies below the 5\% floor and 6 returned the
original unchanged. Low changed 25\% of tokens on average (token edit
distance to the original), High 46\%, and Neutral 10\%.

\paragraph{Pipeline differences.}
\Cref{tab:pipelines} summarizes how the validated pilot differs from
the replays. The replay rewrites have not been rated by humans.

\begin{table}[ht]
\centering
\small
\setlength{\tabcolsep}{3pt}
\begin{tabular}{@{}p{0.26\columnwidth}p{0.33\columnwidth}p{0.35\columnwidth}@{}}
\toprule
 & Range pilot & Replays (both exp.) \\
\midrule
Rewriter & Qwen3.5-27B & DeepSeek Flash \\
Checks & rules, inverse parser, span-aligned fidelity judge & rules, two yes/no questions (GPT Luna 5.6) \\
Retries & up to 2, with reasons & 1, at temperature 0.7 \\
On failure & utterance unavailable; passage dropped & original text delivered \\
Human rating & yes & no \\
\bottomrule
\end{tabular}
\caption{Rewrite pipelines in validation and in the replays.}
\label{tab:pipelines}
\end{table}

\section{Additional Linguistic-Range Validation}

\paragraph{Passage construction.}
The development corpus is three untreated nine-player games (27
players, 280 public utterances). A passage is 150--250 words of
whole utterances by one player in one game, drawn with a fixed seed
from the early, middle, and late thirds of that player's timeline
and kept in speech order. The utterance set is fixed on the
untreated text before any rewrite exists and is identical across
conditions, which yields 24 passages. Each condition is generated
independently from the original. Whole-game transcripts (six per
contrast) form a robustness cohort.

\paragraph{Measures.}
Linguistic features are computed with spaCy and SUBTLEX-US. The
primary measures are word frequency (SUBTLEX Lg10CD), vocabulary
diversity (MATTR, 50-token window), and subordinate clauses per
100 words, with Benjamini--Hochberg correction within this family.
MTLD, content-lemma repetition, dependency distance, parse-tree
depth, noun-phrase length, and sentence length are secondary.
Grammar is checked with LanguageTool errors per 100 words and CoLA
acceptability. Effects average players within each game and weight
games equally; intervals resample games. With three games, the
smallest attainable two-sided $p$ is .25, so the pilot reports
estimates and intervals only.

\begin{table}[ht]
\centering
\small
\setlength{\tabcolsep}{3pt}
\resizebox{\columnwidth}{!}{%
\begin{tabular}{@{}lrrr@{}}
\toprule
Measure & Low$-$Neu. & High$-$Neu. & Neu.$-$Nat. \\
 & ($n{=}15$) & ($n{=}14$) & ($n{=}19$) \\
\midrule
SUBTLEX Lg10CD & $+0.26^{\ddagger}$ & $-0.50$ & $-0.03$ \\
MATTR & $-0.015^{\ddagger}$ & $+0.019$ & $-0.002$ \\
Subord.\ clauses / 100w & $-1.0^{\ddagger}$ & $-0.34$ & $-0.23$ \\
MTLD & $-7.3$ & $+13$ & -- \\
Repetition & $+0.060$ & -- & $-0.018$ \\
Dependency distance & $-0.64$ & $\approx 0$ & -- \\
Tree depth & $-1.35$ & $\approx 0$ & -- \\
Sentence length (words) & $-30$ & $\approx 0$ & $+0.7$ \\
\midrule
LanguageTool errors / 100w & $+0.02$ & -- & $-0.02$ \\
CoLA acceptability & $+0.20$ & -- & $+0.01$ \\
\bottomrule
\end{tabular}}
\caption{Range pilot measures, as passage-level differences between
conditions. $^{\ddagger}$Primary measure whose 95\% game-clustered
interval excludes zero.
$\approx 0$: within noise. --: not reported.}
\label{tab:range-measures}
\end{table}

\paragraph{Sentence-splitting sensitivity.}
Low may lower syntactic measures simply by splitting sentences. On
the 9 passages where Low changed the sentence count least, word
frequency holds ($+0.26$), while MATTR ($-0.007$ [$-0.012$, $+0.001$])
and subordinate clauses ($-0.86$ [$-2.18$, $+0.19$]) keep their
direction with intervals that include zero.

\paragraph{LLM listeners.}
Qwen3.5-27B and GPT Luna 5.6 read two passages by the same player in
both presentation orders. The \emph{range} question asks which passage
shows the more limited range of English, or \emph{same}; a second,
\emph{proficiency} question asks which speaker seems to have the
weaker command of English. Answers are scored $+1$ (treatment judged
more limited), 0 (same), or $-1$ and averaged over orders, listeners,
and passages within a game. On the range question, both listeners
judge Low more limited than Neutral and High, and High less limited
than Neutral, in every passage and both orders (Qwen answers
\emph{same} in 2 of 30 Low--Neutral comparisons). On Neutral versus
Natural, Qwen answers \emph{same} in 97\% of pairs (signed preference
$-0.02$); GPT Luna answers \emph{same} in 63\% and otherwise calls
Natural more limited ($-0.30$ [$-0.57$, $-0.14$]). On the proficiency
question, Qwen calls most pairs the same (70\% of Low--Neutral, 75\%
of High--Neutral) and otherwise picks the expected side
(Low--Neutral $+0.30$, Low--High $+0.71$); GPT Luna picks the
expected side on High--Neutral ($-0.62$) but gives the same answer in
both orders for 7 of 15 Low--Neutral passages ($+0.00$). Range
differences are therefore perceptible, while their reading as weaker
proficiency is less consistent.

\paragraph{Human annotation.}
Two researchers independently labeled every complete pair of the
four comparisons (62 pairs) with the range question, in random order
with randomized sides, on a blind sheet that hid the conditions.
Both finished before adjudication. A third annotator, blind to
condition, resolved the 11 disagreements on a sheet showing only
anonymous item IDs (\Cref{tab:range-validation}).

\paragraph{CEFR ratings.}
As an auxiliary check, Qwen3.5-27B assigned each passage a CEFR band
(A, B, or C) with quoted evidence. Of 18 rated Low passages, 7
received band B and 11 band C; every Neutral, High, and Natural
passage received band C. The rewrites thus move along range without
reaching beginner level, consistent with our construct.

\paragraph{Power.}
From the pilot's between-game variance, six games give 80\% power for
the Low--Neutral contrast on word frequency, MATTR, and the
listener range score, and eight games on subordinate clauses.

\section{Additional Politeness Validation}

\paragraph{Items.}
The 13 source utterances come from 11 untreated seven-player games
and realize \texttt{estimate} (3), \texttt{question} (3),
\texttt{recommend\_action} (4), \texttt{disagree} (1),
\texttt{address\_role} (1), and \texttt{unallied} (1). DeepSeek Flash
rewrote each into all four strategies at temperature 0, using one
prompt per strategy written from Brown and Levinson's definitions
(\Cref{sec:politeness-prompts}). All prompts forbid changing the
speaker, target, polarity, or reasons. Positive politeness may add
rapport but not invented praise or evidence; negative politeness
softens the act, not the truth claim; off-record keeps the target
and proposition recoverable.

\paragraph{Annotation guidelines.}
Each researcher saw the predicate, the original utterance, and the
rewrite, but not the requested strategy or any automatic label. For
each item they gave the dominant strategy (or \emph{ambiguous}), a
confidence, and whether the meaning was preserved (\emph{yes},
\emph{no}, \emph{unsure}). Both researchers are authors of the study.

\begin{table}[ht]
\centering
\small
\setlength{\tabcolsep}{4pt}
\begin{tabular}{@{}lrrrr@{}}
\toprule
Requested & $n$ & Res.\ 1 & Res.\ 2 & Judge \\
\midrule
Bald on-record & 13 & 12 & 13 & 13 \\
Positive & 13 & 12 & 13 & 11 \\
Negative & 13 & 12 & 11 & 13 \\
Off-record & 13 & 9 & 13 & 8 \\
\midrule
All & 52 & 45 & 50 & 45 \\
\bottomrule
\end{tabular}
\caption{Items on which each rater chose the requested strategy, over
all 52 items (including the 3 bald items where the original was
shown).}
\label{tab:pol-recognised}
\end{table}

\paragraph{Inter-annotator agreement.}
The researchers chose the same strategy on 43 of 52 items (83\%,
$\kappa = 0.77$). Four disagreements involve off-record items that
Researcher~1 read as a bald, positive, or negative act, four are
between positive and negative politeness, and one item was labeled
\emph{ambiguous}. On meaning, Researcher~1 answered \emph{yes} on 46
items and Researcher~2 on 37; neither used \emph{unsure}. They agree on
37 items ($\kappa = 0.14$): both say the meaning is kept on 34 and lost
on 3, and 18 items are flagged by at least one. Flags concentrate in
off-record rewrites (12 of 18) and in \texttt{recommend\_action}
(8 of 18), where indirectness can leave the action or target unclear.

\paragraph{Judge.}
The politeness judge is GPT Luna 5.6, run three times per item with
the majority label kept. It sees only the utterance, the speaker, and
the phase, and is never told the requested strategy. It first names
the speech act in at most 12 words and then decides how that act is
realized: stated directly (bald on-record), stated through solidarity
or common ground (positive), stated but softened through deference or
reduced imposition (negative), or left for the listener to infer
(off-record). The prompt instructs it not to label from politeness
markers alone: openers such as ``Interesting that \ldots'', tag
questions, and hedges do not make a stated act off-record. Two further
labels, \emph{ambiguous\_or\_mixed} and \emph{failed\_manipulation},
cover utterances with no dominant strategy and unusable text. The judge
matches Researcher~1 on 42 of 52 items ($\kappa = 0.74$) and
Researcher~2 on 43 ($\kappa = 0.77$). \Cref{tab:judge-confusion}
gives its confusion matrices.

\begin{table}[ht]
\centering
\small
\setlength{\tabcolsep}{4pt}
\begin{tabular}{@{}lrrrrr@{}}
\toprule
Human $\backslash$ Judge & Bald & Pos. & Neg. & Off & Other \\
\midrule
\multicolumn{6}{@{}l}{\emph{Researcher-agreed rewrites ($n=43$)}} \\
Bald     & 12 & 0 & 0 & 0 & 0 \\
Positive & 0 & 11 & 1 & 0 & 0 \\
Negative & 0 & 0 & 10 & 0 & 0 \\
Off      & 1 & 1 & 1 & 6 & 0 \\
\midrule
\multicolumn{6}{@{}l}{\emph{\citet{zhao2025} ($n=135$)}} \\
Bald     & 31 & 0 & 0 & 0 & 0 \\
Positive & 11 & 32 & 0 & 2 & 0 \\
Negative & 10 & 1 & 31 & 1 & 2 \\
Off      & 11 & 0 & 2 & 1 & 0 \\
\bottomrule
\end{tabular}
\caption{Politeness judge against human labels. Other: failed
manipulation or no majority across passes.}
\label{tab:judge-confusion}
\end{table}

On \citet{zhao2025}, of the 90 positive or negative items the judge
placed 64 in that pair, and 63 of those match the published label.
The published off-record items are mostly plain evaluations from
low-rated scenarios (e.g., ``It was not good.''), whose indirectness
comes from the speaker's hidden opinion; the judge reads them as stated
evaluations. This set therefore supports the bald, positive, and
negative categories rather than testing off-record.

\section{Additional Behavioral Analyses}

\paragraph{Replay counts.}
Both experiments replay the same 30 baseline games (seeds
\texttt{pe01}--\texttt{pe30}), which give 289 player-days with a
control. The range experiment runs two control replays and one replay
per level for each player-day ($289 \times 5 = 1{,}445$ replays,
3,615 rewrites). The politeness experiment runs two control replays
and two replays per setting ($289 \times 10 = 2{,}890$ replays,
7,785 treated utterances). Each replay covers one day: discussion,
vote, night, and the next reliability update.

\paragraph{Outcomes.}
Beyond the three outcomes in the main tables, we record the share of
reliability scores that change from the baseline, the share that
change by 2 or more points, and the mean score other players give the
designated speaker (all assessors, and village assessors only).
\Cref{tab:app-range-full,tab:app-pol-full} report them.

\begin{table}[ht]
\centering
\small
\setlength{\tabcolsep}{2.5pt}
\newcommand{\tci}[1]{{\scriptsize #1}}
\resizebox{\columnwidth}{!}{%
\begin{tabular}{@{}lccc@{}}
\toprule
 & Scores moved & Moved 2+ & Score for speaker \\
 & \tci{(pp)} & \tci{(pp)} & \tci{(village, pts)} \\
\midrule
Low $-$ control & +0.24 & +0.70 & +0.03 \\
 & \tci{[$-$0.59, +0.99]} & \tci{[+0.01, +1.43]} & \tci{[$-$0.02, +0.09]} \\
Neutral $-$ control & $-$0.00 & +0.06 & +0.04 \\
 & \tci{[$-$0.87, +0.85]} & \tci{[$-$0.60, +0.76]} & \tci{[$-$0.02, +0.08]} \\
High $-$ control & +0.07 & +0.73 & +0.02 \\
 & \tci{[$-$0.89, +0.93]} & \tci{[+0.13, +1.31]} & \tci{[$-$0.03, +0.07]} \\
Low $-$ Neutral & +0.24 & +0.64 & $-$0.01 \\
 & \tci{[$-$0.60, +1.08]} & \tci{[$-$0.05, +1.37]} & \tci{[$-$0.05, +0.03]} \\
High $-$ Neutral & +0.07 & +0.67 & $-$0.01 \\
 & \tci{[$-$0.73, +0.83]} & \tci{[+0.13, +1.21]} & \tci{[$-$0.07, +0.04]} \\
\bottomrule
\end{tabular}}
\caption{Additional reliability outcomes for linguistic range (30
seeds, 289 player-days), with 95\% seed-clustered intervals.}
\label{tab:app-range-full}
\end{table}

\begin{table}[ht]
\centering
\small
\setlength{\tabcolsep}{2.5pt}
\newcommand{\tci}[1]{{\scriptsize #1}}
\resizebox{\columnwidth}{!}{%
\begin{tabular}{@{}lccc@{}}
\toprule
 & Scores moved & Moved 2+ & Score for speaker \\
 & \tci{(pp)} & \tci{(pp)} & \tci{(village, pts)} \\
\midrule
control level & 30.08\% & 10.03\% & $-$0.11 \\
\midrule
\texttt{all\_bald} & +0.42 & +0.02 & $-$0.00 \\
 & \tci{[$-$0.04, +0.91]} & \tci{[$-$0.45, +0.52]} & \tci{[$-$0.05, +0.05]} \\
\texttt{all\_positive} & $-$0.29 & +0.19 & $-$0.08 \\
 & \tci{[$-$0.87, +0.28]} & \tci{[$-$0.32, +0.70]} & \tci{[$-$0.14, $-$0.02]} \\
\texttt{trust\_positive} & +0.07 & +0.03 & $-$0.03 \\
 & \tci{[$-$0.54, +0.67]} & \tci{[$-$0.44, +0.48]} & \tci{[$-$0.08, +0.02]} \\
\texttt{trust\_bald} & $-$0.04 & $-$0.03 & $-$0.01 \\
 & \tci{[$-$0.64, +0.58]} & \tci{[$-$0.37, +0.29]} & \tci{[$-$0.05, +0.04]} \\
\bottomrule
\end{tabular}}
\caption{Additional reliability outcomes for politeness, each setting
minus the control (30 seeds, 289 player-days).}
\label{tab:app-pol-full}
\end{table}

\paragraph{Sensitivity: excluding \texttt{pe01}.}
Repeating both analyses without seed \texttt{pe01} (29 seeds, 278
player-days) leaves the conclusions unchanged. For range, Low$-$Neutral
on votes is +4.2 pp [+1.0, +7.4] and High$-$Neutral +4.0 pp [+1.2,
+6.8]; trust differences stay within $\pm$0.01 points. For politeness,
\texttt{all\_bald} on skill targets is +2.3 pp [+0.4, +4.2], and
\texttt{all\_positive} lowers the village-assessed score for the
speaker by 0.08 points [$-$0.15, $-$0.02]; every vote interval
includes zero.

\paragraph{Natural-play politeness by phase and speech act.}
In the 30 natural-play games (\Cref{sec:natural-politeness}), werewolves
won 18 and villagers 12. Non-bald speech is most common in the sheriff
election (18\% of utterances) and on Day~2 (15\%), and least common on
Day~1 and Day~3 or later (10\% each). By speech act,
\texttt{recommend\_action} is non-bald in 18\% of cases (positive
politeness 15\%), \texttt{estimate} in 14\% (off-record 9\%),
\texttt{disagree} in 5\%, and \texttt{question} in 1\%. In 1,218 of
1,429 utterances all three judge passes agree; 3 have no majority.
Requiring unanimous labels gives $p = .33$ for the role comparison and
$p = .046$ for the village trust association. Role tests permute roles
among the seven players of each game; trust tests shuffle labels among
utterances by the same speaker, in the same game, with the same speech
act. The role test uses 10,000 permutations; neither test is
corrected for multiple comparisons.

\begin{table}[ht]
\centering
\small
\begin{tabular}{@{}lrr@{}}
\toprule
Listener & $n$ & Non-bald \\
\midrule
\multicolumn{3}{@{}l}{\emph{Village-side speakers: score for listener}} \\
Low (1--3)   & 44  & 2 (5\%) \\
Mid (4--7)   & 481 & 53 (11\%) \\
High (8--10) & 12  & 4 (33\%) \\
\midrule
\multicolumn{3}{@{}l}{\emph{Werewolf speakers}} \\
Teammate (score 10)      & 26  & 2 (8\%) \\
Village player, 4--7     & 299 & 42 (14\%) \\
Village player, 8--9     & 19  & 0 (0\%) \\
\bottomrule
\end{tabular}
\caption{Non-bald utterances in natural play by the speaker's
reliability score for the listener (881 addressed, rated utterances).}
\label{tab:app-natural-trust}
\end{table}

Among werewolves, speech to teammates does not differ detectably from
speech to others ($p = .73$). Pooling both camps weakens the village
association ($p = .16$), because the werewolf score does not measure
trust.

\section{Reproducibility Details}

\paragraph{Models.}
\Cref{tab:models} lists every model and its role. Game agents and
in-game rewriters were called through the DeepSeek API as
\texttt{deepseek-chat}, which the API reported as
\texttt{deepseek-flash}. All game calls use temperature 0 except the
regeneration of a failed rewrite (0.7). The model is not
deterministic at temperature 0, which is why every comparison is made
against a re-sampled control.

\begin{table}[ht]
\centering
\small
\setlength{\tabcolsep}{3pt}
\begin{tabular}{@{}p{0.34\columnwidth}p{0.6\columnwidth}@{}}
\toprule
Model & Role \\
\midrule
DeepSeek Flash (\texttt{deepseek-chat}) & game agents; in-game range and politeness rewriter; politeness validation rewriter \\
GPT Luna 5.6 (\texttt{gpt-5.6-luna}) & in-game gate questions; politeness judge; second range listener \\
Qwen3.5-27B & range pilot rewriter, fidelity judge, inverse parser, first listener, CEFR rater \\
\bottomrule
\end{tabular}
\caption{Models used in this study.}
\label{tab:models}
\end{table}

\paragraph{Seeds and data.}
The replay experiments share baseline seeds \texttt{pe01}--\texttt{pe30};
the natural-play analysis uses a separate set of 30 games
(\texttt{game01}--\texttt{game30}); the politeness validation items
come from the untreated text of 11 of the \texttt{pe} baselines; the range pilot uses three nine-player games.
Each replay records, per treated utterance, the listener, the
assigned and delivered strategy or level, both attempts, the failure
reasons, and the original and delivered text.

\paragraph{Statistics.}
Intervals are percentile intervals from a cluster bootstrap over
seeds (2,000 draws, fixed random seed). Two-sided $p$-values come from
a sign-flip test that flips each seed's mean difference (5,000
permutations). Holm correction is applied across all contrasts and
measures in a table. Human-validation proportions use exact binomial
intervals.

\paragraph{Software and commands.}
The code is written in Python ($\geq$3.10); linguistic measures use
spaCy ($\geq$3.7), SUBTLEX-US, LanguageTool, and a CoLA classifier via
\texttt{transformers}. The main commands are:
\begin{itemize}[leftmargin=*,itemsep=0pt]
\item \texttt{python -m experiments.language\_flip.code.run flip}
and \texttt{\ldots analysis export | effects} (range replays);
\item \texttt{python -m experiments.politeness\_experiment.code.run flip}
and \texttt{\ldots analysis export | effects | judge} (politeness);
\item \texttt{python -m experiments.politeness\_validation.code.analysis}
(politeness validation tables);
\item \texttt{python -m experiments.language\_validation.validation report}
(range pilot measures and listeners).
\end{itemize}

% \paragraph{Cost.}
% The two replay experiments cost about US\$95 in API fees (\$35 for
% range, \$60 for politeness).
% \todo{Add the cost of baseline games, the natural-play games, and
% the validation runs, and the Qwen3.5-27B compute.}

\end{document}